\documentclass[10pt,conference]{IEEEtran}
\IEEEoverridecommandlockouts

\usepackage{cite}
\usepackage{url}
\usepackage{amsmath,amssymb,amsfonts}
\usepackage{makecell}
\usepackage{algorithmic}
\usepackage{graphicx}
\usepackage{textcomp}
\usepackage{float}
\usepackage{xcolor}
\usepackage{booktabs}
\usepackage{multirow}
\usepackage{soul}
\def\BibTeX{{\rm B\kern-.05em{\sc i\kern-.025em b}\kern-.08em
    T\kern-.1667em\lower.7ex\hbox{E}\kern-.125emX}}
\begin{document}

\title{BTGenBot-2: Efficient Behavior Tree Generation with Small Language Models}

\author{
    \IEEEauthorblockN{
        Riccardo Andrea Izzo, 
        Gianluca Bardaro, and
        Matteo Matteucci
    }
    \IEEEauthorblockA{
        Department of Electronics, Information, and Bioengineering\\
        Politecnico di Milano, Milano\\
        (riccardo.izzo, gianluca.bardaro, matteo.matteucci)@polimi.it
    }
}

\maketitle

\begin{abstract}
Recent advances in robot learning increasingly rely on LLM-based task planning, leveraging their ability to bridge natural language with executable actions. While prior works showcased great performances, the widespread adoption of these models in robotics has been challenging as 1) existing methods are often closed-source or computationally intensive, neglecting the actual deployment on real-world physical systems, and 2) there is no universally accepted, plug-and-play representation for robotic task generation. Addressing these challenges, we propose BTGenBot-2, a 1B-parameter open-source small language model that directly converts natural language task descriptions and a list of robot action primitives into executable behavior trees in XML. Unlike prior approaches, BTGenBot-2 enables zero-shot BT generation, error recovery at inference and runtime, while remaining lightweight enough for resource-constrained robots. We further introduce the first standardized benchmark for LLM-based BT generation, covering 52 navigation and manipulation tasks in NVIDIA Isaac Sim. Extensive evaluations demonstrate that BTGenBot-2 consistently outperforms GPT-5, Claude Opus 4.1, and larger open-source models across both functional and non-functional metrics, achieving average success rates of 90.38\% in zero-shot and 98.07\% in one-shot, while delivering up to 16× faster inference compared to the previous BTGenBot.
\end{abstract}


\section{Introduction}
\begin{figure*}[t]
    \centering
    \includegraphics[width=1\linewidth]{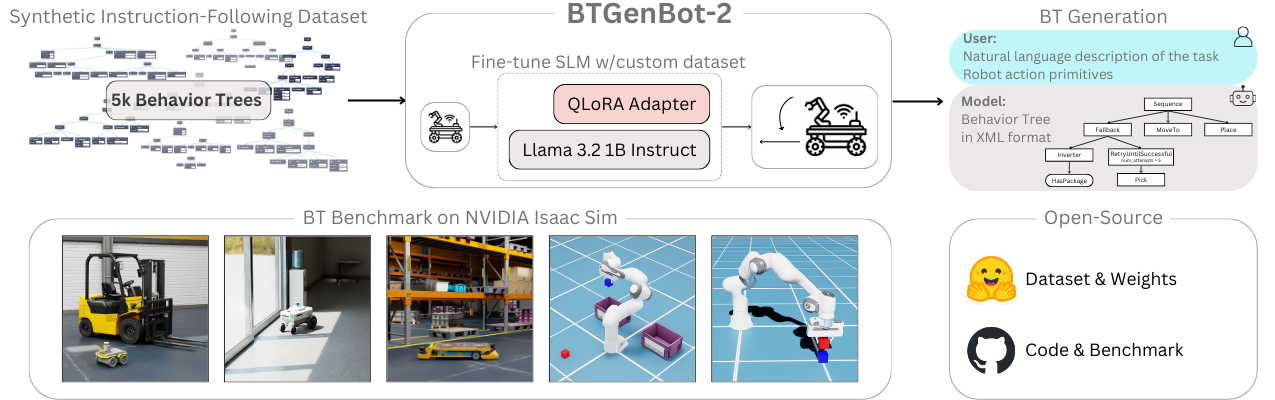}
    \caption{We present \textbf{BTGenBot-2}, a 1B-parameter open-source small language model (SLM), trained on 5k natural language instructions and behavior tree pairs from a synthetic instruction-following dataset. BTGenBot-2 achieves state-of-the-art performance in generating directly executable behavior trees.  We open-source our synthetic dataset, model weights, codebase, and the BT Benchmark to support dissemination and reproducibility.}
    \label{fig:btgenbot-2}
\end{figure*}

The process of autonomously generating task sequences directly executable by robots has sparked interest in the robotic community following the emergence of foundation language models such as GPT-4~\cite{achiam2023gpt}, Gemini~\cite{team2024gemini}, and Llama~\cite{touvron2023llama}. These models have demonstrated impressive reasoning capabilities~\cite{huang2022towards}, being able to represent commonsense knowledge~\cite{zhao2023large}, grounding language to physical objects~\cite{song2023llm}, and adapting plans based on contextual information~\cite{gupta2024action}. Consequently, various representations have been exploited to generate robotics tasks using Large Language Models (LLMs), such as Planning Domain Definition Language (PDDL)~\cite{guan2023leveraging}, Linear Temporal Logic (LTL)~\cite{pan2023data}, Python code~\cite{liang2023code, singh2023progprompt} and Behavior Trees (BTs)~\cite{colledanchise2018behavior, ghzouli2023behavior}. In particular, BTs have gained popularity due to their scalable and structured approach to managing complex robotic tasks, with the BehaviorTree.CPP library becoming a standard component in the Robot Operating System 2 (ROS2) ecosystem, primarily because of its inclusion in the Navigation2 stack~\cite{macenski2020marathon}. 
Despite these advances, most prior works focus on the integration of closed-source models, such as OpenAI's GPT~\cite{achiam2023gpt}, into their pipeline, neglecting the actual deployment on resource-constrained robots and relying on APIs that are not accessible in offline scenarios. Although some studies achieve impressive results, two major limitations prevent their practical adoption: 1) current models are \textit{closed-source}, or with limited transparency into the model architecture, training procedures, and data mixture, and 2) the dependency on computationally intensive models that cannot be feasibly executed on actual robotic hardware. We argue that advancing robotics research requires the development of open-source and efficient models building on open-source LLMs~\cite{touvron2023llama, team2024gemma}. This enables the released model to be deployed locally on real robots without relying on external APIs or additional components. Additionally, the field still lacks standardized benchmarks for evaluating LLM-driven BT generation in robotics.
To this end, we introduce BTGenBot-2, a 1B parameter open-source Small Language Model (SLM) that establishes a new state-of-the-art in generating executable and ROS2-compatible BTs. As illustrated in Figure \ref{fig:btgenbot-2}, BTGenBot-2 is built upon a pretrained SLM, further fine-tuned with compute-efficient methods~\cite{hu2022lora, dettmers2023qlora} on a custom synthetic instruction-following dataset. Unlike the previous approach, BTGenBot-2 enables zero-shot BT generation, is resilient to errors at both inference and runtime, and is compact enough to run on consumer-grade GPUs while preserving state-of-the-art performance. We further introduce a comprehensive benchmark of 52 tasks spanning navigation and manipulation, organized into three levels of difficulty, and implemented in NVIDIA Isaac Sim\footnote{\url{https://developer.nvidia.com/isaac/sim}}. Our experimental results show that BTGenBot-2 outperforms BTGenBot~\cite{izzo2024btgenbot}, a 7B-parameter state-of-the-art model for BT generation, and proprietary models such as OpenAI's GPT-5 and Anthropic's Claude Opus 4.1 across functional and non-functional metrics. Our main contributions are the following: 
\begin{enumerate}
    \item We release a new synthetic instruction-following dataset of 5,204 pairs of executable BTs and natural language task descriptions, filling a gap in planning datasets by providing a foundation for generalisable BT generation.
    \item We introduce BTGenBot-2, a fully open-source SLM for on-device BT generation within the ROS2 stack
    \item We design two novel failure detection mechanisms to handle errors at both inference and runtime, significantly enhancing robustness.
    \item We propose the first standardized benchmark for evaluating LLM-based BT generation, enabling reproducible and comparable assessment across functional and non-functional metrics. BTGenBot-2 is validated extensively with this benchmark in simulation and on real robots, establishing itself as a strong BT generator.
\end{enumerate}
We publicly release our model, dataset, benchmark, and deployment code\footnote{\url{https://airlab-polimi.github.io/BTGenBot-2/}}.

\section{Related Work}
\subsection{Behavior Trees (BTs)}
A BT is formally defined as a directed rooted tree where the internal nodes are called \textit{control flow nodes} and the leaf nodes are called \textit{execution nodes}~\cite{colledanchise2018behavior}. The root node has no parent, while all other nodes form a parent-child relationship. During execution, a periodic tick signal originates at the root node and propagates through the tree from left to right. Nodes execute only when ticked and return one of the possible status signals: \textit{Success}, \textit{Failure}, or \textit{Running}. The classical representation of a BT comprises four categories of control flow nodes (Sequence, Fallback, Parallel, and Decorator) and two categories of execution nodes (Action and Condition). Prior studies comparing BTs to other representations such as Finite State Machines (FSMs), PDDL, and Decision Trees highlight the advantages of BTs in terms of modularity, reactivity, and interpretability~\cite{biggar2021expressiveness}. These benefits, in addition to the availability of libraries such as BehaviorTree.CPP\footnote{\url{https://www.behaviortree.dev/}} and pytrees\footnote{\url{https://py-trees.readthedocs.io/en/devel/}}, have further driven the utilization.

\subsection{LLMs in robotics}
Large Language Models (LLMs) represent a class of deep learning models typically built upon the Transformer architecture. Although they were originally tailored for text generation~\cite{brown2020language}, LLMs have since demonstrated remarkable capabilities in instruction-following~\cite{zhang2023instruction}, decision-making~\cite{li2022pre}, reasoning~\cite{huang2022towards} and in-context learning~\cite{brown2020language}. Nowadays, among the most widely recognized LLMs, there are GPT~\cite{achiam2023gpt}, Gemini~\cite{team2024gemini} and Llama~\cite{touvron2023llama}. Their rapid adoption has also extended to the field of robotics, where LLMs have been leveraged for language-based task generation~\cite{brohan2023can, liang2023code, izzo2024btgenbot}, end-to-end action generation with visual representation~\cite{zitkovich2023rt, kim2024openvla}, and human-robot interaction~\cite{kim2024understanding}.

\subsection{BTs generation using LLMs} 
Recent efforts in robotics increasingly leverage LLMs or foundation models to achieve high-level task planning. Early methods translate user commands into Python scripts, mapping language-based instructions to code primitives~\cite{liang2023code, singh2023progprompt, brohan2023can}, or into classical planning languages such as PDDL~\cite{Zhou2023ISRLLMIS} and LTL~\cite{pan2023data}. One emerging approach revolves around the direct generation of BTs from a natural-language description of the task~\cite{lykov2024llm, izzo2024btgenbot, ao2024llm, Chen2024IntegratingIU, zhou2024llm, styrud2024automatic}. LLM-BrAIn~\cite{lykov2024llm} uses completely synthetic data and relies only on human evaluation, neglecting practical robot deployment. BTGenBot~\cite{izzo2024btgenbot} leverages real-world tested BTs to fine-tune a lightweight model, offers a comprehensive evaluation framework, and outputs XML-based BTs compatible with the BehaviorTree.CPP library. LLM-as-BT-Planner~\cite{ao2024llm} presents four in-context learning strategies, including a human-in-the-loop approach, to build BTs from user directives. OBTEA~\cite{Chen2024IntegratingIU} encodes goals as well-formed first-order logic formulas and processes these via an LLM before expanding the plan into a BT. 
Similarly, LLM-BT~\cite{zhou2024llm} incorporates semantic maps showcasing robustness and adaptability to environmental changes, while BETR-XP-LLM~\cite{styrud2024automatic} employs dynamic expansion to resolve errors in both planning and execution phases. Prior works~\cite{ao2024llm, Chen2024IntegratingIU, zhou2024llm, styrud2024automatic} leverage closed-source GPT models or rely on external APIs, making them unsuitable for deployment on a real robot. Our method utilizes an open-source SLM directly deployable on a robot and can operate on consumer hardware. Furthermore, while~\cite{ao2024llm, Chen2024IntegratingIU} heavily rely on prompting techniques such as in-context learning and few-shot demonstrations, our model performs zero-shot without requiring the user to provide any additional examples. In addition,~\cite{Chen2024IntegratingIU, styrud2024automatic} focus on the expansion of pre-existing BTs, while our approach directly generates executable BTs. Finally, aside from BTGenBot \cite{izzo2024btgenbot} and our framework, existing approaches to BT generation are closed-source and not ROS2–compatible. Consequently, we will consider BTGenBot as our main baseline.
    
\section{Methodology}
\begin{figure*}[t]
    \centering
    \includegraphics[width=0.9\linewidth]{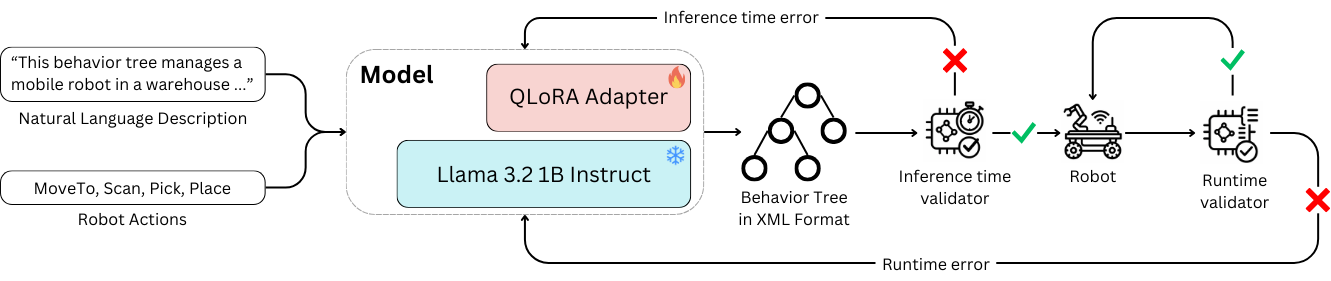}
       \caption{\textbf{Overview of the model architecture.} The model takes as input a natural language task description of a robotic task along with the set of available robot action primitives, generating a ROS2-compatible BT in XML format. The model is adapted using a QLoRA adapter while keeping pre-trained parameters frozen. The generated BTs are initially validated at inference time, checking for syntax and action-space consistency before execution. Additionally, at runtime, an inline logger tracks stack traces and blackboard states, triggering subtree regeneration in case of errors.}
    \label{fig:pipeline}
\end{figure*}

This section details our methodology for generating executable BTs using an SLM, as outlined in Figure \ref{fig:pipeline}. Building on BTGenBot, we surveyed the top-performing open-source SLMs available on \textit{Open LLM Leaderboard} by Hugging Face and selected \textit{Llama-3.2-1B-Instruct}~\cite{grattafiori2024llama}, an instruction-tuned model with an 8k context length. Section \ref{sec:problem_formulation} formulates the problem. Section \ref{sec:dataset} details our data curation pipeline, while Section \ref{sec:fine-tuning} describes the QLoRA fine-tuning process. Finally, Section \ref{sec:error-handling} outlines inference and runtime error recovery mechanisms, while Section \ref{sec:benchmark} introduces our standardized framework for evaluating LLM-based BT generation.

\begin{figure}[t]
    \centering
    \includegraphics[width=0.9\linewidth]{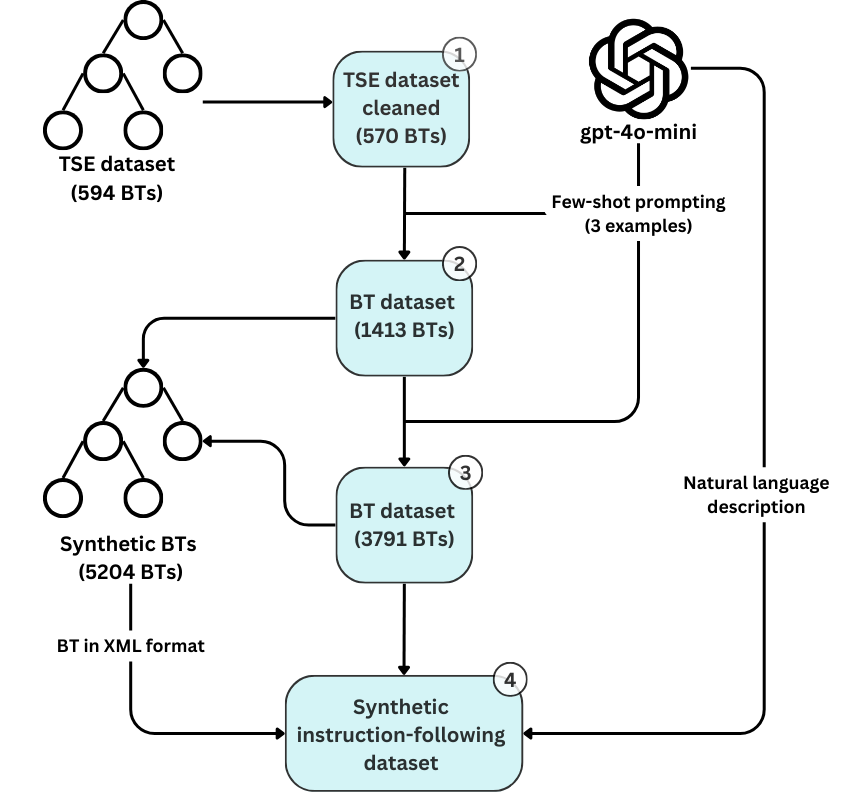}
    \caption{\textbf{Dataset generation.} Starting with the TSE dataset, a new instruction-following dataset is created through four key steps: (1) cleanse the raw XML data, (2) for each original BT, use \textit{gpt‑4o‑mini} to generate three variants, (3) repeat step 2 with the new dataset, (4) merge all resulting datasets while producing a natural-language description for each BT.}
    \label{fig:dataset_generation}
\end{figure}

\subsection{Problem Formulation}
\label{sec:problem_formulation}
We formulate the generation of BTs as a supervised learning task. Let $\mathcal{D} = \{(\mathcal{I}_i, \mathcal{A}_i, \mathcal{T}_i)\}_{i=1}^N$ be a dataset where $\mathcal{I}$ represents a natural language instruction, $\mathcal{A}$ is the set of robot action primitives, and $\mathcal{T}$ is the corresponding behavior tree in XML format. Our objective is to fine-tune a SLM parameterized by $\theta$ to maximize the probability of generating the correct behavior tree $\mathcal{T}$ given the inputs:
\begin{equation}
\theta^* = \operatorname*{argmax}_\theta \sum_{(\mathcal{I}, \mathcal{A}, \mathcal{T}) \in \mathcal{D}} \log P_\theta(\mathcal{T} \mid \mathcal{I}, \mathcal{A})
\end{equation}
The generated output is implicitly constrained to be a valid XML schema compatible with BehaviorTree.CPP, containing only leaf nodes present in the action space $\mathcal{A}$.

\begin{figure}[t]
    \centering
    \includegraphics[width=0.9\linewidth]{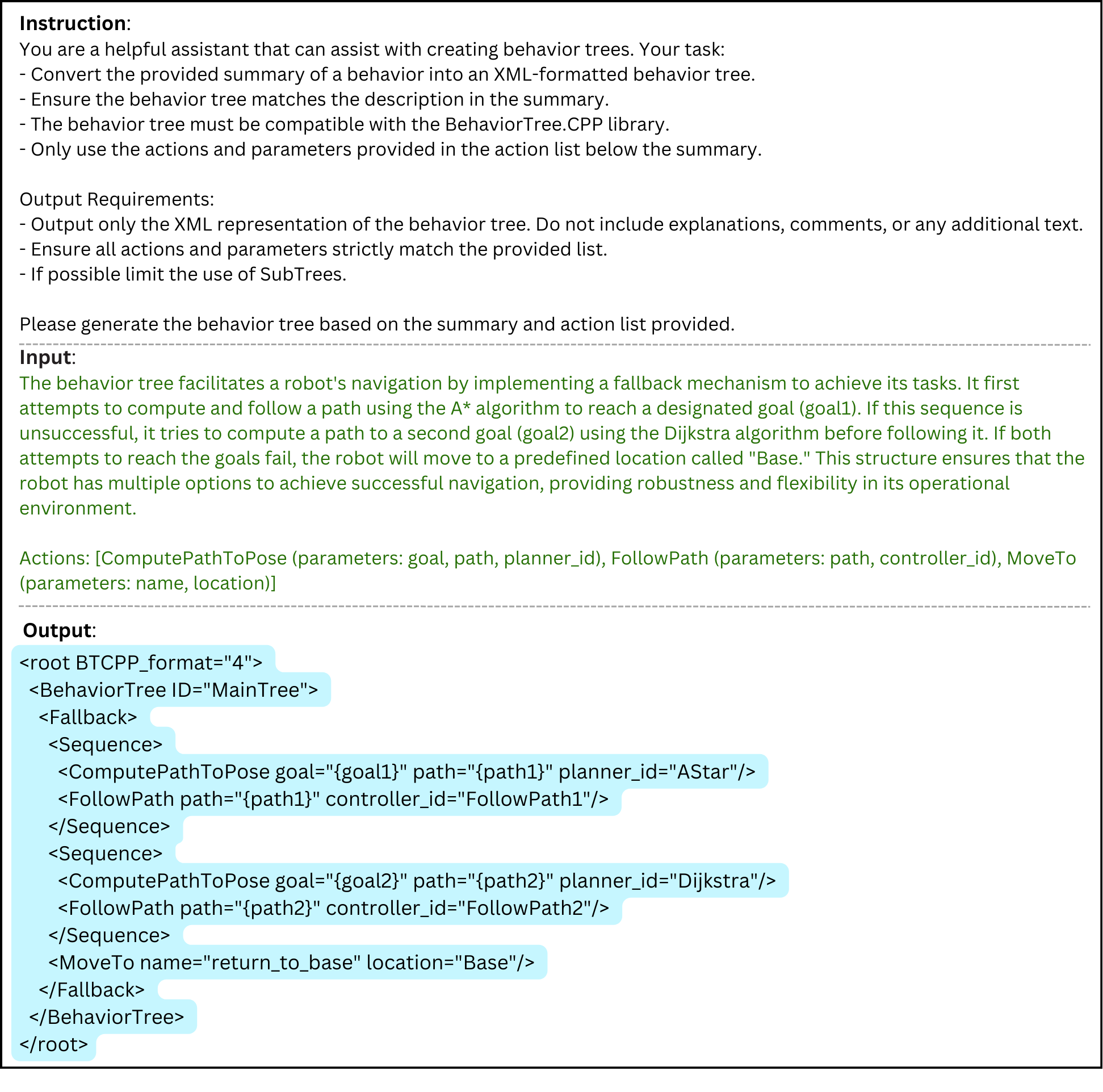}
    \caption{\textbf{Dataset sample.} A representative example from the generated instruction-following dataset with its three components: the \textit{instruction} that provides system contextual information, the \textit{input} comprising a natural-language task description and its corresponding robot actions, and the \textit{output} showcasing the generated XML-based behavior tree.}
    \label{fig:dataset_sample}
\end{figure}

\subsection{Dataset}
\label{sec:dataset}
We created a new instruction-following dataset following the Alpaca schema~\cite{taori2023alpaca}, which includes three components for each sample: \textit{instruction}, \textit{input}, and \textit{output}. The \textit{instruction} component provides system contextual information for the model, including its role, the required output format, and any additional constraints. The \textit{input} component is a natural-language description of a robotic task paired with the list of available robot actions. Finally, the \textit{output} component is an XML-based BT that fits the provided description of the task. The foundation of our dataset is the TSE dataset~\cite{ghzouli2023behavior}, which includes approximately 600 behavior trees sourced from open-source robotics projects. These BTs have been validated on physical robots, ensuring high-quality and reliable real-world implementations. As illustrated in Figure \ref{fig:dataset_generation}, the generation of our dataset is articulated in four steps:

\textbf{(1) Data Preparation.} We start with 594 BTs from the TSE dataset~\cite{ghzouli2023behavior}. We first verified syntactic correctness and compatibility with \textit{BehaviorTree.CPP} using a Python XML parser. After these validation steps, we are left with 570 BTs.

\textbf{(2) First BT Generation.} From this curated set, we used \textit{gpt-4o-mini} to produce three new variants of each BT, following the approach of~\cite{taori2023alpaca}, which generates instruction-response pairs from a curated human-written instruction-response pairs seed set. We chose \textit{gpt-4o-mini} due to its fast and low-cost inference. Then, inspired by~\cite{honovich2022unnatural}, we employed nucleus sampling (top-p = 0.99) to encourage creativity while using the same few-shot examples, but we constrained the generation so that the execution nodes remain unchanged. This ensured compatibility with the underlying robotic APIs and  \textit{BehaviorTree.CPP}. Through a final cleaning phase, this process produced 1413 synthetic BTs.

\textbf{(3) Second BT Generation.} We repeated the process from step~(2), with the newly generated 1413 BTs as the input set, obtaining an additional 3791 synthetic BTs. This step further increased the overall diversity of the dataset.

\textbf{(4) Instruction Generation.} We merged the two synthetic BT datasets from steps~(2-3), resulting in a total of 5204 BTs. For each of these BTs, \textit{gpt-4o-mini} then generates: (i) a brief natural language description of the task (less than 200 words), and (ii) a list of its corresponding action nodes and parameters. This step completes the dataset by pairing each BT with a task description and the required robot actions. 

To evaluate the quality of outputs generated by \textit{gpt-4o-mini}, we conducted a visual assessment on a subset of results. This involved checking for semantic coherence between the natural language descriptions and their corresponding BTs, as well as verifying the syntactic correctness of the BTs using an XML parser. 
Figure \ref{fig:dataset_sample} shows a representative example from the resulting dataset. 
Although generating additional samples might have further diversified the data, our preliminary experiments suggested decreasing returns and BT quality decline. The resulting dataset reflects real-world scale by preserving the original training distribution in terms of task types. The BTs average 11.23 nodes and 10.23 transitions, yet extend to up to 157 nodes and 156 transitions, covering both common BTs and larger ones used in complex industrial settings.
Our dataset provides a significant advantage over existing BT collections. For instance,~\cite{izzo2024btgenbot} offers a high-quality real-world dataset, but the number of training tokens is significantly below what would be needed for fine-tuning a 7-billion-parameter model. Conversely, the dataset introduced by~\cite{lykov2024llm} is purely synthetic. While the size of approximately 8500 samples is closer to the recommended token counts, it lacks the same level of real-world validation as the BTs sourced from the TSE dataset. Our dataset combines the strengths of both approaches: a core of curated real-world BTs, enhanced with controlled synthetic generation to expand size and variety.

\subsection{Fine-Tuning}
\label{sec:fine-tuning}
LLMs fine-tuning is a computationally intensive task, therefore, we employed a Parameter-Efficient Fine-Tuning (PEFT) approach, specifically QLoRA~\cite{dettmers2023qlora}, a quantized extension of LoRA~\cite{hu2022lora}. QLoRA retains LoRA’s benefits, such as avoiding full model retraining, typically training on the order of one-thousandth as many parameters as the original model. For instance, in a 1-billion-parameter model such as \textit{Llama-3.2-1B-Instruct}, LoRA may only require training approximately one million parameters. 15 tokens per parameter heuristic suggests about 15 million tokens for a 1-billion-parameter model~\cite{hoffmann2022training}, whereas our dataset amounts to about 3.5 million tokens, an order of magnitude below the target. Despite falling short of the previous heuristics, we decided to proceed based on emerging evidence that even relatively small but high-quality datasets can effectively fine-tune LLMs to achieve competitive performance~\cite{bansal2024smaller}. For training, we used the dataset described in Section \ref{sec:dataset}, which was partitioned into training and test sets with a conventional 95\%-5\% split. Most of the QLoRA hyperparameters were left at default (e.g., \textit{lora\_r}, \textit{lora\_alpha}, and \textit{lora\_dropout}). Following~\cite{dettmers2023qlora}, we expanded the original set of LoRA target modules beyond the attention layers ['k\_proj', 'q\_proj', 'v\_proj', 'o\_proj'] to include the MLP layers ["gate\_proj", "down\_proj", "up\_proj"]. Typical LLMs train for at most one or two epochs, but we found that using up to five epochs was beneficial for our comparatively smaller dataset, yielding training accuracies exceeding 95\%. Regarding the learning rate, we kept the standard fixed value of 1e-4. We used a learning rate warmup with a ratio of 0.1 to improve training stability and reduce early overfitting. The final model was trained for approximately 30 hours on two NVIDIA RTX Quadro 6000 with a total of 48GB of VRAM, using a batch size of 16. 

\subsection{Error Handling and Recovery}
\label{sec:error-handling}
A key motivation for adopting BTs as our representation is their inherent ability to handle failures and enable recovery strategies. In BTGenBot-2, we explicitly address two sources of errors that must be considered when automatically generating BTs: inference time errors that occur when generating BTs from natural language, and runtime errors that arise during execution on a robot.

\subsubsection{Inference Time Validator}
\label{sec:inference-time-error}
LLMs often produce outputs that may be syntactically malformed or semantically inconsistent with the robot’s action space. To ensure deployability, BTGenBot-2 incorporates a validation layer that constrains decoding to XML format and enforces a closed vocabulary of BT nodes and action primitives. Any malformed output is automatically rejected, leading to a new generation. This validation layer takes advantage of a Python parser, designed for BTs, using a YAML file that describes all the allowed primitives, thus avoiding the risk of hallucination when generating the BT.
This layered validation ensures that, before execution, every BT produced by BTGenBot-2 is both syntactically correct and aligned with the robot primitives of the target robotic system, eliminating the need for human intervention.

\subsubsection{Runtime Validator}
Unlike flat action sequences, BTs support reactive execution through control flow nodes such as Fallback and Retry. Additionally, BTGenBot-2 focuses on local recovery mechanisms that can be encoded directly in the generated BT. This is achieved with a C++ in-process logger and implemented by every tree node. When any leaf or condition node returns "Failure", the node records the stack trace and a snapshot of the blackboard. If a local recovery (e.g., Fallback, Retry) succeeds, no regeneration is prompted. Otherwise, if the failure is expected to propagate, the runtime parser requests a subtree regeneration from the LLM. Upon validation with the inference time validator, the original tree is modified with the new sub-tree. This mechanism further enhances reactivity and robustness, enabling the system to correct even unexpected errors during execution.

\subsection{Behavior Tree Benchmark}
\label{sec:benchmark}
Evaluating the generation of BTs in robotics is inherently challenging as it requires evaluating both structural validity and execution. However, existing approaches often rely on ambiguous representations, unsupported by widely adopted frameworks such as ROS2 BehaviorTree.CPP and py\_trees, and on narrowly defined task suites, limiting the possibility of reproducible comparison. Additionally, effective testing of articulated BTs requires a complex but deliberately diverse robotic setup to ensure comprehensive coverage of BT variants and task domains. With LLMs now able to synthesize BTs from natural language, the absence of a common and reproducible benchmark has become a critical bottleneck. We propose a standardized benchmark for evaluating BT generation, adopting \textit{BehaviorTree.CPP} as the canonical syntax, given its compatibility with ROS2. The benchmark comprises 52 tasks across navigation and manipulation and is organized into three difficulty levels. Navigation includes 32 tasks, with 12 easy, 10 medium, and 10 hard. Manipulation includes 20 tasks, with 6 easy, 8 medium, and 6 hard. Overall, the benchmark contains 18 easy, 18 medium, and 16 hard tasks. Navigation tasks include point-to-point and waypoint-sequence problems at the easy level, priority and conditional navigation at the medium level, and mixtures of these at the hard level, with particular emphasis on the use of novel action primitives, fault recovery and more complex reasoning. Manipulation tasks focus on tabletop scenarios, including pick-and-place, stacking, sorting, and object reordering. To stress generalization, the hard tasks are intentionally out-of-distribution relative to the original BTs in our dataset, providing a substantially harder test of a model’s ability to serve as a broadly generalizable BT generator. All tasks are implemented in \textit{NVIDIA Isaac Sim}, a widely used simulation environment for robot learning. 

\section{Experiments and Results}
\label{sec:experimental_results}
The goal of our experimental evaluation is to assess the performance and effectiveness of our model as a powerful tool to generate robotic tasks through the use of BTs and natural language instructions. Specifically, we benchmark our model against prior works that employ LLMs for BT generation, and against current state-of-the-art generalist LLMs. In our evaluation, BTGenBot~\cite{izzo2024btgenbot} serves as the primary baseline for LLM-based BT generation, while GPT-5 and Claude Opus 4.1 represent the leading general-purpose LLMs.

\begin{figure}[t]
    \centering
    \includegraphics[width=0.7\linewidth]{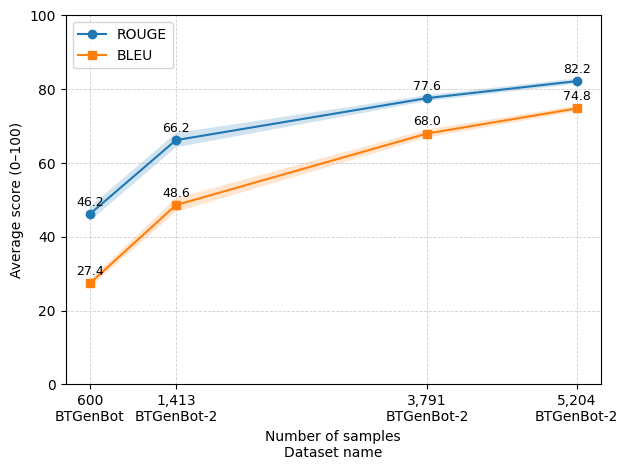}
    \caption{\textbf{Data scaling evaluation.} Average ROUGE and BLEU scores (mean ± std) with increasing dataset size: ROUGE: 46.2 ± 1.94, 66.2 ± 1.94, 77.6 ± 0.80, 82.2 ± 0.75; BLEU: 27.4 ± 1.02, 48.6 ± 1.85, 68.0 ± 1.10, 74.8 ± 0.75, corresponding to 600, 1,413, 3,791, and 5,204 samples. Standard deviation evaluated across 5 runs with temperature=0.9}
    \label{fig:data_scaling}
\end{figure}

\begin{figure*}[t]
    \centering
    \includegraphics[width=1\linewidth]{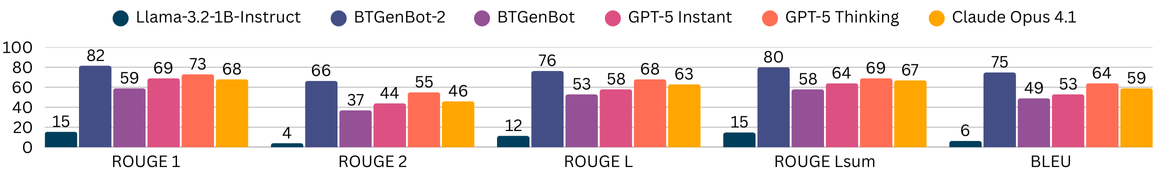}
    \caption{\textbf{ROUGE and BLEU Scores.} Evaluation on a test set of 250 BTs from the custom instruction-following dataset demonstrates that fine-tuning substantially improves performance compared to the pre-trained baselines, with BTGenBot-2 achieving the highest scores. GPT-5 Thinking performs competitively, with GPT-5 Instant and Claude Opus 4.1 falling slightly behind. The original BTGenBot is outscored by its successor, BTGenBot-2.}
    \label{fig:rouge-bleu-score}
\end{figure*}

\subsection{Data Curation Pipeline Effectiveness}
\label{sec:data_curation_effectiveness}
In this section, we assess how dataset scale affects BT generation quality, thereby validating the curation pipeline introduced in Section~\ref{sec:dataset}. We fine-tune \textit{Llama-3.2-1B-Instruct} on four dataset configurations with the procedure described in Section \ref{sec:fine-tuning}, varying the number of training epochs according to dataset size while keeping other hyperparameters fixed. To assess the model's convergence toward the target data distribution, we employed text-based metrics such as ROUGE~\cite{lin-2004-rouge} and BLEU~\cite{papineni2002bleu}. The ROUGE score measures the similarity between the generated text and the reference text using overlapping n-grams and word sequences that appear in both texts. The BLEU score measures the overall quality of text with respect to the original. All models were evaluated on a test split comprising 5\% of each dataset configuration. As shown in Figure \ref{fig:data_scaling}, both metrics increase monotonically with additional training samples. The largest gain occurs when scaling from the prior BTGenBot corpus to a curated subset of 1,413 BTs. Further scaling up yields smaller but consistent improvements, leading to convergence. These results provide evidence that our data curation pipeline, with a curated corpus of BTs, practically improves BT generation.

\subsection{Preliminary Evaluation Phase}
\label{sec:preliminary_evaluation}
In this preliminary evaluation phase, we benchmarked six models for BT generation quality on a test set of 250 BTs spanning navigation and manipulation tasks. The \textit{Llama-3.2-1B-Instruct} model was used as a lower-bound reference, while its fine-tuned variant, \textit{BTGenBot-2}, represents our proposed approach. As baselines, we considered \textit{BTGenBot}, as well as leading proprietary LLMs such as OpenAI \textit{GPT-5} (Instant and Thinking modes) and Anthropic \textit{Claude Opus 4.1}. Model quality was initially assessed using ROUGE 1/2/L/Lsum and BLEU. While these scores cannot strictly evaluate the semantic validity of a BT, they serve as an essential preliminary validation for assessing the model's ability to acquire the correct XML syntax and action usage. ROUGE 1/2 captures unigram and bigram overlap (lexical recall and word-order sensitivity), while ROUGE L/Lsum evaluates sequence-level structural similarity through the longest common subsequence. As shown in Figure \ref{fig:rouge-bleu-score}, \textit{BTGenBot-2} achieves the best scores across all metrics, outperforming even the strongest proprietary baseline (\textit{GPT-5 Thinking}). In contrast, the base pre-trained version of the model exhibited the worst performance, highlighting the importance of domain-specific fine-tuning. Intermediate performance was observed for \textit{BTGenBot} and \textit{GPT-5 Instant}, with the latter slightly outperforming the original \textit{BTGenBot}.

\subsection{Validation Phase}
\label{sec:validation}

\begin{table*}[t]
\centering
\caption{Evaluation of models on Functional and Non-Functional metrics}
\resizebox{\textwidth}{!}{%
\begin{tabular}{l ccc ccc ccc c | ccc}
\toprule
 & \multicolumn{10}{c|}{\textbf{Functional}} 
 & \multicolumn{3}{c}{\textbf{Non-Functional}} \\
\cmidrule(lr){2-11} \cmidrule(lr){12-14}
\textbf{Model} 
& \textbf{SR (E)} & \textbf{P@3 (E)} & \textbf{Time (E)}
& \textbf{SR (M)} & \textbf{P@3 (M)} & \textbf{Time (M)} 
& \textbf{SR (H)} & \textbf{P@3 (H)} & \textbf{Time (H)} 
& \textbf{Avg SR} 
& \textbf{Action Coherency} & \textbf{XML Syntax} & \textbf{Semantic Correctness} \\
\midrule
\multicolumn{14}{c}{\textbf{Zero-shot}} \\
\midrule
BTGenBot-2-ER     & \textbf{18 / 18} & \textbf{18 / 18} & 5.8s & \textbf{17 / 18} & \textbf{17 / 18} & 8.2s & \textbf{12 / 16} & \textbf{13 / 16} & 11.7s & \textbf{90.38\%} & \textbf{100\%}  & \textbf{100\%}  & \textbf{94.23\%}  \\
BTGenBot-2      & 17 / 18 & \textbf{18 / 18} & \textbf{5.1s} & 15 / 18 & 16 / 18 & \textbf{7.5s} & \textbf{12 / 16} & 12 / 16 & \textbf{11.0s} & 84.61\% & \textbf{100\%}  & 96.15\%  & \textbf{94.23\%}  \\
BTGenBot               & 13 / 18 & 15 / 18 & 85s & 10 / 18 & 12 / 18 & 98s & 7 / 16 & 8 / 16 & 178s & 57.69\% & 63.46\%  & 78.84\%  & 46.15\%   \\
GPT-5 Instant          & 15 / 18 & 16 / 18 & 12.8s & 14 / 18 & 15 / 18 & 15.0s & 5 / 16 & 7 / 16 & 15.2s & 65.38\% & 88.46\%  & 82.69\%  & 80.76\%  \\
GPT-5 Thinking         & 17 / 18 & \textbf{18 / 18} & 34.2s & 15 / 18 & 15 / 18 & 36.8s & 5 / 16 & 6 / 16 & 39.7s & 71.15\% & 92.30\%  & 86.53\%  & 86.53\%  \\
Claude Opus 4.1-ER & \textbf{18 / 18} & \textbf{18 / 18} & 23.9s & 15 / 18 & 15 / 18 & 28.1s & 7 / 16 & 9 / 16 & 30.4s & 76.92\% & 92.30\% & 90.38\%  & 88.46\%  \\
Claude Opus 4.1 & 17 / 18 & \textbf{18 / 18} & 23.1s & 13 / 18 & 14 / 18 & 27.4s & 4 / 16 & 6 / 16 & 29.8s & 65.38\% &  90.38\% & 82.69\%  & 80.76\%  \\
\midrule
\multicolumn{14}{c}{\textbf{One-shot}} \\
\midrule
BTGenBot-2-ER      & \textbf{18 / 18} & \textbf{18 / 18} & \textbf{8.4s} & \textbf{18 / 18} & \textbf{18 / 18} & 9.6s & \textbf{15 / 16} & \textbf{15 / 16} & 13.1s & \textbf{98.07\%} & \textbf{100\%}  & \textbf{100\%}  & \textbf{98.07\%}  \\
BTGenBot-2     & \textbf{18 / 18} & \textbf{18 / 18} & \textbf{8.4s} & 16 / 18 & 17 / 18 & \textbf{8.9s} & 14 / 16 & \textbf{15 / 16} & \textbf{12.6s} & 92.38\% & \textbf{100\%}  & 98.07\%  & 94.23\%  \\
BTGenBot               & 16 / 18 & 17 / 18 & 101s & 14 / 18 & 14 / 18 & 114s & 7 / 16 & 8 / 16 & 197s & 71.15\% & 73.07\%  & 84.61\%  & 51.92\%  \\
GPT-5 Instant          & \textbf{18 / 18} & \textbf{18 / 18} & 15.3s & 16 / 18 & 16 / 18 & 16.2s & 7 / 16 & 8 / 16 & 18.8s & 78.84\% & 96.15\%  & 86.53\%  & 84.61\%  \\
GPT-5 Thinking         & \textbf{18 / 18} & \textbf{18 / 18} & 35.8s & 17 / 18 & 17 / 18 & 36.8s & 9 / 16 & 10 / 16 & 42.2s & 84.61\% & \textbf{100\%}  & 92.30\% & 88.46\%  \\
Claude Opus 4.1-ER & \textbf{18 / 18} & \textbf{18 / 18} & 25.5s & \textbf{18 / 18} & \textbf{18 / 18} & 29.8s & 10 / 16 & 12 / 16 & 34.3s & 88.46\% & \textbf{100\%} & 96.15\% & 94.23\%  \\
Claude Opus 4.1 & \textbf{18 / 18} & \textbf{18 / 18} & 25.0s & \textbf{18 / 18} & \textbf{18 / 18} & 29.4s & 8 / 16 & 10 / 16 & 33.6s & 84.61\% & 96.15\%  & 90.38\%   & 88.46\%  \\
\bottomrule
\end{tabular}%
\label{tab:evaluation}
}
\end{table*}

In this validation phase, we evaluate the same models as in Section \ref{sec:preliminary_evaluation}, excluding the base version \textit{Llama-3.2-1B-Instruct} due to its consistently low performance in Section \ref{sec:preliminary_evaluation}. Our model, BTGenBot-2, is designed to work zero-shot, promoting a plug-and-play system that is intuitive and accessible to end users. This eliminates the need for prompt engineering or the inclusion of examples, while preserving high-quality BT generation. However, for a fair comparison, we also evaluated the model in a one-shot setting, allowing direct comparison with proprietary models that were not fine-tuned on our dataset. For BTGenBot-2, we consider an additional version, namely BTGenBot-2-ER (Error Recovery), which employs the two validators outlined in Section \ref{sec:error-handling}. To evaluate the generalizability of our ER module, we conducted an ablation study by applying it also to Claude Opus 4.1. The goal of this phase is to assess the end-to-end pipeline as a generalist BT generator. Evaluation is performed on the benchmark proposed in Section \ref{sec:benchmark}, with results reported separately for each level. For the evaluation, we use a mixture of functional and non-functional metrics~\cite{gugliermo2024evaluating}. The functional metrics assess whether a BT executes its intended functions:
\begin{itemize}
    \item \textbf{Success Rate (SR)}: a BT is successful if it is executable and achieves the goal state.
    \item \textbf{Pass@k (P@k)}: measures the likelihood that at least one of the top-k generated candidates is correct~\cite{chen2021evaluating}, where correctness is determined by the definition of SR.
    \item \textbf{Inference Time}: time required by the 4-bit quantized model to generate a complete BT, measured on an NVIDIA GTX 1080 GPU (8 GB VRAM).
\end{itemize}
Complementary, non-functional metrics measure aspects of the BT that are not directly related to its functionality and apply regardless of its performance:
\begin{itemize}
    \item \textbf{Action Coherency}: verifies that the BT uses only the predefined robotic actions from the task prompt, computed with the Python parser in Section \ref{sec:inference-time-error}.
    \item \textbf{XML Syntax}: assess the correctness of the XML schema and the compatibility with the BehaviorTree.CPP library, using the system of Section \ref{sec:inference-time-error}.
    \item \textbf{Semantic Correctness}: evaluated by three human experts using a binary scale to measure whether the structure and logic of the BT align with the natural language task description. Final decisions were made by majority vote.
\end{itemize}

Table \ref{tab:evaluation} compares the models on easy/medium/hard tasks in both zero-shot and one-shot settings, using both functional and non-functional metrics. All models were prompted uniformly, receiving only the task description and the list of available robot primitives. In the zero-shot setting, BTGenBot-2 achieves an average success rate of 84.61\%, outscoring the strongest baseline GPT-5 Thinking by almost 14\%, and surpassing by a larger margin GPT-5 Instant and Claude Opus 4.1, as well as the previous BTGenBot by more than 30\%. As expected, Pass@3 improves absolute accuracy without altering the relative ranking of models. In terms of efficiency, BTGenBot-2 generates a BT in 11 seconds on average, comparable only to GPT-5 Instant, and up to 3× faster than other proprietary reasoning models and 16× faster than the original BTGenBot. Regarding non-functional metrics, BTGenBot-2 demonstrates strong reliability: it is the only model that consistently uses only the provided robot actions, while the original BTGenBot consistently fails due to the original fine-tuning dataset not including the list of actions in the input prompt. XML syntax correctness is nearly perfect in BTGenBot-2, outperforming other models that occasionally produce invalid outputs. BTGenBot-2 demonstrates good capabilities in generating BTs that are syntactically correct and fully compatible with BehaviorTree.CPP. Concerning semantic correctness, our model captures task intent most accurately, followed by GPT-5 Thinking, while GPT-5 Instant, Claude Opus 4.1, and BTGenBot lag behind. With one-shot prompting, each model receives a single example of BT generation. All models benefit from the additional example, with proprietary models averaging around 80\% success rate. BTGenBot-2 achieves the best performance at 92.38\%, while also remaining the fastest model despite longer average inference times due to the more complex one-shot prompt. For non-functional metrics, only BTGenBot-2 and GPT-5 Thinking achieve perfect action coherency, though GPT-5 Thinking lags with the other metrics. Considering the variant with error recovery, BTGenBot-2-ER, the integration of the two validators further improves performance in both zero-shot and one-shot settings, achieving an average success rate of 90.38\% and 98.07\%, respectively. Notably, in nearly all cases, a single iteration is sufficient to successfully correct the BT. This improvement comes at the negligible cost of less than one additional second of inference time on average, due to the regeneration of incorrect BTs. XML syntax reaches 100\%, ensuring completely valid and executable BTs. Results in both zero-shot and one-shot scenarios demonstrate that the error recovery module is beneficial even when applied to generalist LLMs such as Claude Opus 4.1. Notably, the ER module increases the model's success rate, particularly in hard tasks, while significantly improving all the non-functional metrics, achieving perfect action coherency and high syntax correctness. Overall, the results show that BTGenBot-2-ER delivers the strongest combination of reliability, structural fidelity, and efficiency across tasks and prompting settings. It achieves substantial gains over the previous BTGenBot and consistently outperforms proprietary models, establishing itself as a robust and effective tool for generating complex BTs from natural language instructions.

\subsection{Real Robot Validation}
\label{sec:real_robot_validation}
We assessed the practical applicability of our pipeline with real-world trials on physical robots. For navigation, we used an AgileX Scout, a mobile wheeled robot with skid-steering kinematics, while for manipulation we employed an SO-ARM 101, a 6-DoF manipulator. Within this setup, we tested the benchmark’s easy tasks and achieved a success rate of $17/18$ with Pass@3, improving to $18/18$ with error recovery. These experiments demonstrate that BTGenBot-2 can be effectively deployed in real-world scenarios. Since the benchmark includes complex tasks that extend beyond the constraints of our laboratory setup, the results in Table \ref{tab:evaluation} provide a more comprehensive evaluation of the model's capabilities.

\section{Conclusion}
We presented BTGenBot-2, a 1B-parameter open-source SLM that translates natural language task descriptions into executable BTs. Through a carefully curated synthetic dataset, compute-efficient fine-tuning, and two novel validators, BTGenBot-2 delivers strong performance while remaining lightweight enough for deployment on resource-constrained robots. We also released the first standardized benchmark for LLM-based BT generation, spanning 52 navigation and manipulation tasks at varying difficulty levels, providing a reproducible basis for evaluation. Experiments in NVIDIA Isaac Sim and qualitative tests on a real robot show that BTGenBot-2 consistently outperforms prior systems, including its predecessor and large foundation models such as GPT-5 and Claude Opus 4.1. These results highlight the advantages of SLM for robotic task planning, bridging the gap between a natural language instruction of the task and a practical representation that is executable on a real robot. 

\bibliographystyle{IEEEtran}
\bibliography{bibliography}

\end{document}